\documentclass{article}
\usepackage{ijcai26}

\usepackage{times}
\usepackage{soul}
\usepackage{url}
\usepackage[hidelinks]{hyperref}
\usepackage[utf8]{inputenc}
\usepackage[small]{caption}
\usepackage{graphicx}
\usepackage{amsmath}
\usepackage{amsthm}
\usepackage{booktabs}
\usepackage{algorithm}
\usepackage[switch]{lineno}

\usepackage{subcaption}
\usepackage{acronym}
\usepackage{amssymb}
\usepackage{mathtools}
\usepackage{algpseudocode} 

\title{Decoupling Return-to-Go for Efficient Decision Transformer}

\author{
Yongyi Wang$^1$
\and
Hanyu Liu$^1$\and
Lingfeng Li$^1$\and
Bozhou Chen$^1$\and
Ang Li$^1$\and
Qirui Zheng$^1$\and
Xionghui Yang$^1$\and
Wenxin Li$^{1,\ast}$
\affiliations
$^1$School of Computer Science, Peking University, Beijing, China.\\
\emails
$^\ast$lwx@pku.edu.cn
}

\begin{document}
\acrodef{rl}[RL]{Reinforcement Learning}
\acrodef{dt}[DT]{Decision Transformer}
\acrodef{rtg}[RTG]{Return-to-Go}
\acrodef{ddt}[DDT]{Decoupled DT}
\acrodef{adaln}[adaLN]{Adaptive Layer Normalization}
\acrodef{dp}[DP]{Dynamic Programming}
\acrodef{mdp}[MDP]{Markov Decision Process}
\acrodef{pomdp}[POMDP]{Partially Observable MDP}
\acrodef{ln}[LN]{Layer Normalization}
\acrodef{dit}[DiT]{Diffusion Transformer}
\acrodef{mlp}[MLP]{Multi-Layer Perceptron}
\acrodef{mse}[MSE]{Mean Squared Error}
\acrodef{bc}[BC]{Behavior Cloning}
\acrodef{sota}[SOTA]{state-of-the-art}
\acrodef{him}[HIM]{Hindsight Information Matching}
\acrodef{rcsl}[RCSL]{Reward Conditioned Supervised Learning}
\acrodef{gcsl}[GCSL]{Goal Conditioned Supervised Learning}
\acrodef{llm}[LLM]{Large Language Model}

\newcommand{\pr}{P}
\newcommand{\real}{\mathbb{R}}
\maketitle

\begin{abstract}
The Decision Transformer (DT) has established a powerful sequence modeling approach to offline reinforcement learning. It conditions its action predictions on Return-to-Go (RTG), using it both to distinguish trajectory quality during training and to guide action generation at inference. In this work, we identify a critical redundancy in this design: feeding the entire sequence of RTGs into the Transformer is theoretically unnecessary, as only the most recent RTG affects action prediction. We show that this redundancy can impair DT's performance through experiments. To resolve this, we propose the Decoupled DT (DDT). DDT simplifies the architecture by processing only observation and action sequences through the Transformer, using the latest RTG to guide the action prediction. This streamlined approach not only improves performance but also reduces computational cost. 
Our experiments show that DDT significantly outperforms DT and establishes competitive performance against state-of-the-art DT variants across multiple offline RL tasks.
\end{abstract}

\section{Introduction}
\label{sec:intro}
Offline \ac{rl} aims to learn effective policies from static, pre-collected datasets, eliminating the need for costly or risky online interaction during training. 
This paradigm is especially valuable for real-world applications that face the dual challenges of scarce data and high safety demands, including robotic manipulation and autonomous driving.
However, offline \ac{rl} faces fundamental challenges, such as overcoming the distributional shift between the dataset and the learned policy, and effectively extracting high-quality behavioral patterns from potentially suboptimal data.

\ac{dt} \cite{chen2021decision} has emerged as a novel and promising paradigm that departs from traditional approaches based on offline \ac{dp}. 
The \ac{dt} leverages the Transformer \cite{vaswani2017attention} architecture to model trajectories as sequences of (\ac{rtg}, observation, action) tokens, predicting each action autoregressively based on past observations, actions, and \ac{rtg}s. 
This sequence modeling paradigm circumvents the Markov assumption and the deadly triad problem \cite{peng2024deadly}, delivering robust performance across offline \ac{rl} benchmarks.

Central to \ac{dt} is the use of \ac{rtg}, which specifies the cumulative reward desired from a given timestep onward. 
During training, the \ac{rtg} sequence serves as a crucial feature to distinguish trajectory quality.
During inference, it acts as an adjustable, user-specified target to guide the action generation. 
The standard \ac{dt} design follows an intuitive autoregressive scheme: it feeds the entire sequence of \ac{rtg}s, observations, and actions into the Transformer to predict the next action. 
In this work, we challenge this established design. 

Through a combination of theoretical analysis and empirical evidence, we identify a critical redundancy in the standard \ac{dt} implementation: conditioning the Transformer on the entire historical sequence of \ac{rtg}s is theoretically unnecessary for optimal action prediction under the trajectory modeling objective, as only the most recent \ac{rtg} provides essential future information.
More importantly, our experiments reveal that this redundancy is not benign; it can introduce unnecessary complexity, thereby degrading the model's computational efficiency and final performance, a particularly undesirable trait for applications such as robotic manipulation.

\begin{figure}[t] 
    \centering
    \includegraphics[width=1.0\linewidth]{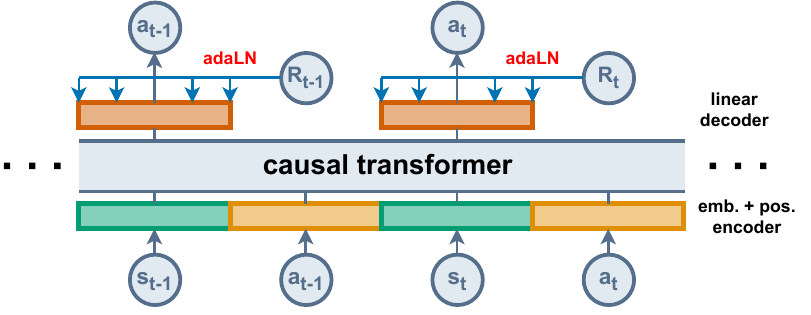}
    \caption{DDT architecture. States/observations and actions are first projected via modality-specific linear embeddings, followed by a positional encoding. A GPT backbone with causal self-attention processes the resulting tokens. RTGs are integrated via adaLN to modulate the output token for action prediction.}
    \label{fig:ddt}
\end{figure}

To address this limitation, we propose a simple but highly effective architecture: \ac{ddt}\footnote{Code in the supplementary, will be online upon acceptance}. 
As demonstrated in Fig. \ref{fig:ddt}, the core insight is to decouple the conditioning of \ac{rtg} from the Transformer's input sequence. 
Specifically, \ac{ddt} takes only the sequences of observations and actions as input to the Transformer. 
The conditioning is simplified: only the last \ac{rtg} is used to directly modulate the output token corresponding to the predicted action via \ac{adaln} \cite{peebles2023scalable}. 
This elegant design drastically reduces input redundancy, allowing the Transformer to focus on learning essential dynamics and policy patterns from observation-action histories, while remaining goal-directed via a focused \ac{rtg} signal.

Extensive experiments on widely-adopted offline \ac{rl} D4RL benchmarks \cite{fu2020d4rl} demonstrate that \ac{ddt} not only consistently outperforms the original \ac{dt} and its variants, but also achieves competitive results compared to advanced model-free and model-based offline algorithms \ac{rl}. 
Furthermore, by eliminating the need to process lengthy \ac{rtg} sequences, \ac{ddt} enjoys reduced computational overhead and a smaller memory footprint, making it a more efficient and scalable solution, a property that is advantageous for deployment in computationally cost-aware real-world scenarios.

Our principal contributions are:
\begin{itemize}
    \item Exposing the \textbf{fundamental redundancy} of conditioning on the full \ac{rtg} history in \ac{dt};
    \item Introducing the \ac{ddt}, a \textbf{simplified architecture} that conditions only on the current \ac{rtg} with \ac{adaln};
    \item Empirically showing that \ac{ddt} is competitive with state-of-the-art \ac{dt} variants on D4RL datasets.
\end{itemize}

The remainder of this paper is organized as follows: 
Section \ref{sec:relwork} reviews related work. 
Section \ref{sec:prelim} provides necessary background. %
Section \ref{sec:theory} details our theoretical motivation. %
Section \ref{sec:ddt} demonstrates our proposed \ac{ddt} architecture. %
Section \ref{sec:exp} presents experimental results and analysis. %
Section \ref{sec:discuss} explores possible extensions of the \ac{ddt} framework.
Finally, Section \ref{sec:conclude} concludes the paper.

\section{Related Work}
\label{sec:relwork}
\subsection{Offline \ac{rl}}
Offline \ac{rl} learns policies from static datasets, offering a safe and cost-effective paradigm for applications like robotics and autonomous driving \cite{levine2020offline}. 
Predominant methods in offline RL are built upon the framework of \ac{dp} and value function learning. However, learning from a static dataset introduces the core challenge of distributional shift. To address this, several key algorithmic families have been developed, including policy constraint methods (e.g., BCQ \cite{fujimoto2019off}), value regularization methods (e.g., CQL \cite{kumar2020conservative}), and expectile-regression-based approaches (e.g., IQL \cite{kostrikov2021offline}). Despite their success, these methods, due to their reliance on bootstrapping within the \ac{dp} framework, remain susceptible to error accumulation and sensitivity to hyperparameters.

\subsection{Offline RL via Sequence Modeling}
Departing from offline \ac{rl} algorithms based on \ac{dp}, the \ac{dt} \cite{chen2021decision} frames \ac{rl} as sequence modeling. 
By leveraging a Transformer to autoregressively predict actions conditioned on past observations, actions, and \ac{rtg}s, \ac{dt} avoids bootstrapping and the Markov assumption, establishing a robust, trajectory modeling paradigm. 
Generalized \ac{dt} \cite{furutageneralized} later provided a unifying theoretical framework of \ac{him}.
Subsequent research has targeted specific limitations of \ac{dt}.
\emph{Suboptimal Trajectory Stitching}: Enhanced by generating waypoints (WT \cite{badrinath2023waypoint}), adaptive context lengths (EDT \cite{wu2023elastic}), or Q-learning relabeling (QDT \cite{yamagata2023q}).
\emph{Stochasticity \& Over-optimism}: Addressed by clustering returns (ESPER \cite{paster2022you}) or disentangling latent variables (DoC \cite{yangdichotomy}), advantage conditioning (ACT \cite{gao2024act}).
\emph{Online Training \& Adaptation}: Explored via online fine-tuning (ODT \cite{zheng2022online}), auxiliary critics (CGDT \cite{wang2024critic}), \ac{rl} gradient infusion (TD3+ODT \cite{yan2024reinforcement}), value function guidance (VDT \cite{zheng2025value}).
Models other than Transformer have also been explored, including convolutional models (DC \cite{kim2023decision}), Mamba \cite{gu2024mamba} (Decision Mamba \cite{lv2024decision,huang2024decision}), and hybrid designs (LSDT \cite{wang2025long}).

\subsection{Conditioning via Adaptive Normalization}
\ac{ln} \cite{ba2016layer} was introduced to address the issue of internal covariate shift in recurrent neural networks, thereby stabilizing the training process.
This layer-wise normalization proved particularly effective for sequential and Transformer-based architectures, where batch statistics are often unstable. 
Building upon \ac{ln} and inspired by conditional normalization techniques from earlier generative models \cite{de2017modulating,karras2019style,dhariwal2021diffusion}, the \ac{dit} \cite{peebles2023scalable} formally established and popularized the modern \ac{adaln} module.
In this design, a shared \ac{mlp} maps conditioning vectors, such as timestep or class embeddings, into layer-specific modulation parameters (scale $\gamma$ and bias $\beta$). 
This mechanism dynamically fuses conditional information into the network, effectively replacing the fixed, learnable affine transformations traditionally used in each \ac{ln} layer of a Transformer block with adaptive, condition-dependent scaling and shifting.

\section{Preliminaries}
\label{sec:prelim}
\subsection{MDP and POMDP}
In \ac{rl}, the environment is commonly formulated as a \ac{mdp}, denoted by $\mathcal{M}=\langle S,A,T,r,\rho\rangle$, where $S$ is the state space, $A$ is the action space, $T:S\times A\to\Delta S$ is the transition probability, $r:S\times A\to\real$ is the reward function, $\rho\in\Delta S$ is the initial state distribution.

A \ac{pomdp} generalizes the \ac{mdp} framework to partially observable environments, denoted as $\mathcal{N} = \langle S, A, T, r, \rho, \Omega, O \rangle$, where $\langle S,A,T,r,\rho\rangle$ is an \ac{mdp}, $\Omega$ is the set of observations, and $O:S\times A\to\Delta\Omega$ is the observation probability function.

The objective in \ac{mdp} and \ac{pomdp} is to learn a policy that maximizes the cumulative return $R_1=\sum_{t=1}^T r_t$.
For an \ac{mdp}, a policy specifies a distribution over actions given a state. 
In a \ac{pomdp}, however, the state is unobservable, necessitating the extraction of a belief state \cite{aastrom1965optimal,smallwood1973optimal,subramanian2022approximate} $b\in\Delta S$ from the trajectory to inform decisions, where
\begin{equation}
\label{eq:bs}
b_{t+1}(s_{t+1})=\frac{O(o_t\mid s_{t+1},a_t)\sum_{s_t\in S}T(s_{t+1}\mid s_t,a_t)b_t(s_t)}{\pr(o_t\mid a_t,b_t)}
\end{equation}

It is noteworthy that reward information is not utilized in computing the belief state in Eq. \ref{eq:bs}, as per the standard assumption in a \ac{pomdp}. 
The belief state serves as a sufficient statistic \cite{kaelbling1998planning}, implying that optimal decisions can be made without other information within the trajectory when it is given.
\subsection{Decision Transformer}
The \ac{dt} is trained with a \ac{mse} \ac{bc} loss to output an action given a trajectory as context. 
At step $t$, the model takes as input a trajectory $$\tau_t = (\hat{R}_{t-k+1}, o_{t-k+1}, a_{t-k+1}, ..., \hat{R}_{t-1}, o_{t-1}, a_{t-1}, \hat{R}_t, o_t)$$ where $k$ denotes the specified context length and $\hat{R}_t = \sum_{t'=t}^T r_{t'}$ is the \ac{rtg} at step $t$. 
Thus, the \ac{dt} policy takes the form 
\begin{equation}
\label{eq:dt}
    \pi(a_t \mid \tau_t) = \pi(a_t \mid \hat{R}_{t-k+1:t}, o_{t-k+1:t}, a_{t-k+1:t-1})
\end{equation} where the action $a_t$ is conditioned on the sequence of \ac{rtg}s, observations, and actions within the context window of length $k$.
Such a sequence modeling paradigm of \ac{dt} makes it well-suited for solving \ac{pomdp}s as well as \ac{mdp}s.

\subsection{Adaptive Layer Normalization}
AdaLN \cite{peebles2023scalable} injects conditional information by dynamically adjusting the \ac{ln} affine parameters, replacing static $\gamma,\beta$ with learned functions $\gamma(z),\beta(z)$ of the condition vector $z$.
The calculation process of \ac{adaln} is as follows:
\begin{equation}
\label{eq:adaln}
\begin{aligned}
    \gamma(z), \beta(z)&=\text{MLP}(z) \\
    \text{adaLN}(x,z)&=\gamma(z)\odot\frac{x-\mu(x)}{\sqrt{\sigma^2(x)+\epsilon}}+\beta(z)
\end{aligned}
\end{equation}
where $\mu$ is the mean value of each dimension of $x$, $\sigma^2$ is the variance, and $\odot$ denote the Hadamard product.

The number of layers of the \ac{mlp} in \ac{adaln} scales with the complexity of the condition $z$, though in practice it is kept small; a single layer suffices for many cases.

\section{Redundancy of RTG Sequence as Condition}
\label{sec:theory}
This section presents the theoretical rationale for the redundancy of conditioning on the entire \ac{rtg} sequence in the original architecture of \ac{dt}.
In brief, in \ac{dt} policy Eq. \ref{eq:dt}, the set of conditioning random variables $(\mathbf{R}_{t-k+1:t}, \mathbf{O}_{t-k+1:t}, \mathbf{A}_{t-k+1:t-1})$ can be grouped into two parts: the \textbf{trajectory history} $\mathbf{H}_t\coloneqq(\mathbf{O}_{t-k+1:t}, \mathbf{A}_{t-k+1:t-1})$ which refers to the events that have already occurred and been observed during the reasoning process, including executed actions and received observations and the \textbf{future condition} $\mathbf{R}_{t-k+1:t}$. which primarily aims to guide the current action through anticipated future returns. 

Let $\mathbf{r}_t$ denote the random variable representing the reward obtained at time $t$. 
Then the random event $\{\mathbf{R}_{t-k+1:t}=R_{t-k+1:t}\}$ can be expressed as follows:
\begin{equation}
\label{eq:rtgs}
    \begin{aligned}
    &\{\mathbf{R}_{t-k+1:t}=R_{t-k+1:t}\} = \bigcap_{i=t-k+1}^t\{\mathbf{R}_i=R_i\} \\
    &= \{\mathbf{R}_t=R_t\}\cap\bigcap_{i=t-k+1}^{t-1}\{\mathbf{R}_{i+1}-\mathbf{R}_i=R_{i+1}-R_i\} \\
    &= \{\mathbf{R}_t=R_t\}\cap\{\mathbf{r}_{t-k+1:t-1}=r_{t-k+1:t-1}\}
    \end{aligned}
\end{equation}

\begin{figure}[t] 
    \centering
    \includegraphics[width=0.8\linewidth]{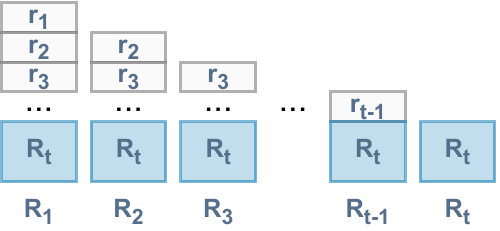}
    \caption{
    In the \ac{rtg} sequence, the difference between successive terms is the reward. These rewards do not contribute to the belief state. Therefore, they are redundant information in history.
    }
    \label{fig:rtg}
\end{figure}

As shown in Eq.~\ref{eq:rtgs} and illustrated in Fig. \ref{fig:rtg}, the future condition of \ac{dt} comprises the intersection of two terms: $\{\mathbf{R}_t=R_t\}$, representing the expected future returns after step $t$, and $\{\mathbf{r}_{t-k+1:t-1}=r_{t-k+1:t-1}\}$, which corresponds to already observed past rewards and therefore contains no information about the future.
In \ac{pomdp}, since the belief state has already extracted all decision-relevant information from the history and, according to Eq. \ref{eq:bs}, is independent of the rewards in the history, $\{\mathbf{r}_{t-k+1:t-1}=r_{t-k+1:t-1}\}$ cannot provide any additional information for decision-making. Therefore, $R_{t-k+1:t-1}$ is redundant information.

Based on the above theoretical results, we conclude that under the standard \ac{pomdp} framework \cite{kaelbling1998planning}, the policy formulation of \ac{dt} contains redundant \ac{rtg} conditioning and should be simplified to:
\begin{equation}
\label{eq:pi}
\pi(a_t\mid\tau_t)=\pi(a_t\mid\hat{R}_t,o_{t-k+1:t},a_{t-k+1:t-1})
\end{equation}
which takes only the latest \ac{rtg} $\hat{R}_t$ as future condition.
\section{DDT: A Simplified Architecture}
\label{sec:ddt}
Based on the conclusions in Section \ref{sec:theory}, significant redundancy exists when conditioning the policy of the original \ac{dt} on \ac{rtg} sequence. 
Therefore, the required new architecture must correspond to a policy of the form given in Eq. \ref{eq:pi}, one that can leverage the powerful sequence modeling capabilities of the Transformer while effectively harnessing the guidance of the latest \ac{rtg}. 
This section details the implementation of our proposed \ac{ddt} framework.

\begin{figure}[t] 
    \centering
    \includegraphics[width=1.0\linewidth]{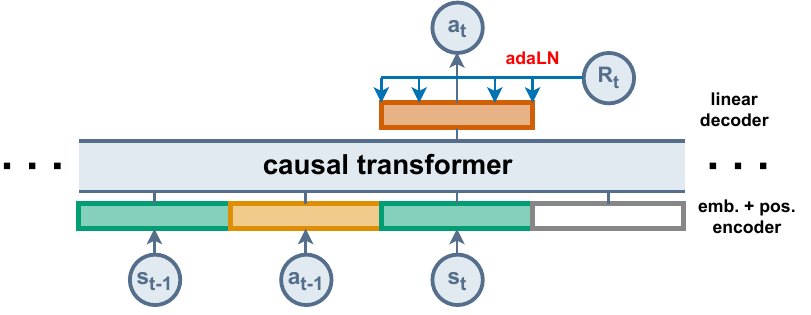}
    \caption{During action prediction, DDT takes only the most recent RTG $\hat{R}_t$ as condition. This condition applies to the hidden state corresponding to $o_t$ via adaLN, and the result is subsequently fed into $\text{Pred}_A$ (an MLP) to output the action $a_t$.}
    \label{fig:ddtpred}
\end{figure}

Alg. \ref{alg:ddtpred} and Fig. \ref{fig:ddtpred} outline the action prediction process of \ac{ddt}. 
Its key distinctions from original \ac{dt} lie in: 
\begin{itemize}
    \item Line \ref{algline:mktraj}, where the input sequence to the Transformer is constructed solely from observations and actions, excluding the \ac{rtg} sequence;
    \item Lines \ref{algline:gethidden}–\ref{algline:cond}, where the \ac{rtg} condition $\hat{R}_t$ modulate only the hidden state corresponding to the last action $a_t$.
\end{itemize}

Thus, the design of \ac{ddt} circumvents the need of \ac{dt} to rely on the entire \ac{rtg} sequence, $\hat{R}_{t-k+1:t}$, as input to the Transformer during action prediction. 
With \ac{adaln} implemented as a single linear layer adding negligible overhead, DDT cuts computation by reducing the input sequence length from $3k$ to $2k$, leveraging the quadratic scaling of Transformer inference cost.

\begin{algorithm}[tb]
    \caption{Decoupled Decision Transformer Prediction}
    \label{alg:ddtpred}
    \textbf{Input}: Obs: $o_{t-k+1:t}$, Act: $a_{t-k+1:t-1}$, RTG: $\hat{R}_t$.\\
    \textbf{Parameter}: Transformer, adaLN, $\text{Embed}_O$, $\text{Embed}_A$, $\text{Pred}_A$.\\
    \textbf{Output}: Predicted action $a_t$.

    \begin{algorithmic}[1] 
        \Statex \Comment{\emph{MLP \& Embeddings per-timestep.}}
        \State $o^\prime\gets\text{Embed}_O(o_{t-k+1:t}, t)$
        \State $a^\prime\gets\text{Embed}_A(a_{t-k+1:t-1},t)$
        \Statex \Comment{\emph{Interleave tokens as: $(o_1,a_1,o_2,a_2,\dots)$.}}
        \State $\text{Trajectory}\gets\text{stack}(o^\prime,a^\prime)$ \label{algline:mktraj}
        \State $\text{HiddenStates}\gets\text{Transformer}(\text{Trajectory})$
        \Statex \Comment{\emph{Get the hidden state of the last action $a_t$.}}
        \State $\text{ActionHidden}\gets\text{unstack}(\text{HiddenStates}).\text{actions}[-1]$ \label{algline:gethidden}
        \Statex \Comment{\emph{Incorporate RTG conditioning $\hat{R}_t$.}}
        \State $\text{ConditionedHidden}\gets\text{adaLN}(\text{ActionHidden},\hat{R}_t)$ \label{algline:cond}
        \State \Return $\text{Pred}_A(\text{ConditionedHidden})$
    \end{algorithmic}
\end{algorithm}

As shown in Fig \ref{fig:ddt}, during training, the forward pass of \ac{ddt} differs from its action prediction pipeline (Fig. \ref{fig:ddtpred}, Alg. \ref{alg:ddtpred}): at line \ref{algline:gethidden}, hidden states are extracted for the entire action sequence $a_{t-k+1:t}$ rather than only for the last action $a_t$; and at line \ref{algline:cond}, the complete \ac{rtg} sequence $\hat{R}_{t-k+1:t}$ is used to modulate the corresponding hidden states via \ac{adaln}. 
Our \ac{ddt} design preserves the same forward pass and training loop as the original \ac{dt}. 
The loss function also remains identical as \ac{dt}'s, i.e., the \ac{mse} loss for \ac{bc}.

\section{Experiments}
\label{sec:exp}
This section evaluates the performance of our proposed simplified \ac{dt} framework, \ac{ddt}, on the widely adopted D4RL \cite{fu2020d4rl} benchmark for continuous action space control problems. 
The results show that \ac{ddt} not only reduces the input sequence length and consequently the inference cost compared to the original \ac{dt}, but also substantially surpasses its performance. 
Remarkably, \ac{ddt} remains competitive with recently introduced state-of-the-art \ac{dt} variants that enhance performance by integrating value-based offline \ac{rl} techniques or using more complex architectures.

In addition to the main results presented in Tab. \ref{tab:mainexp}, we investigate the following questions:

\textbf{What explains DDT's superior performance over DT?}

To address this, we visualize the attention scores of \ac{dt} and \ac{ddt} in Section \ref{subsec:attention}, which reveals that \ac{ddt}'s attention distribution aligns more closely with the Markovian nature of the problem. We therefore infer that by removing the redundant \ac{rtg} sequence, \ac{ddt} enables a more efficient allocation of attention scores.

\textbf{Is the adaLN module necessary for achieving the policy form in Eq. \ref{eq:pi}, or can simply modifying the attention mask to block irrelevant RTGs suffice?}

We examine this in Section \ref{subsec:maskout} by comparing DDT to a "blocked-DT" variant, which only modifies the attention mask. While blocked-DT shows a minor improvement over the original DT, its performance remains significantly lower than that of DDT. This demonstrates that using adaLN to integrate the return condition is a more effective approach.

\textbf{Can DDT be applied to tasks with discrete action spaces, high stochasticity, and sparse rewards?}

We validate this in Section \ref{subsec:2048} through experiments in the 2048 game. The results confirm that DDT generalizes effectively to such tasks without any performance loss compared to the original DT.

\textbf{Does adding more linear layers to adaLN help?}

In Section \ref{subsec:addlayer}, we examine this question. The results demonstrate that for the experimental tasks, a single-layer, activation-free adaLN is sufficient for conditioning. In fact, adding more layers and introducing nonlinearity degrades performance.

\subsubsection{Datasets}
\begin{figure}[t] 
    \centering
    \begin{subfigure}[b]{0.48\columnwidth}
        \centering
        \includegraphics[width=\textwidth]{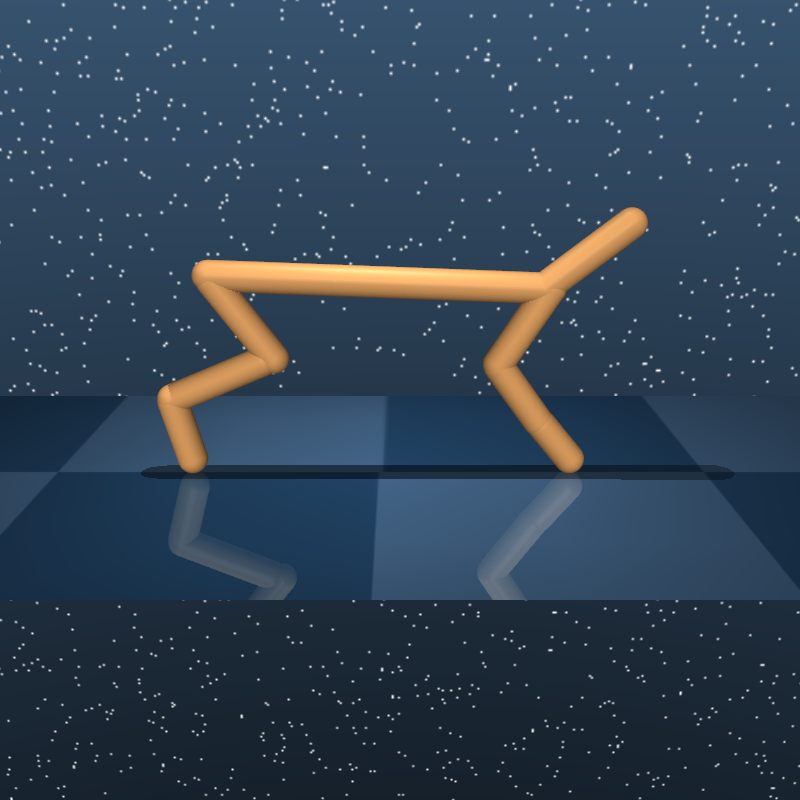}
        \caption{D4RL-HalfCheetah}
        \label{fig:cheetah}
    \end{subfigure}
    \hfill 
    \begin{subfigure}[b]{0.48\columnwidth}
        \centering
        \includegraphics[width=\textwidth]{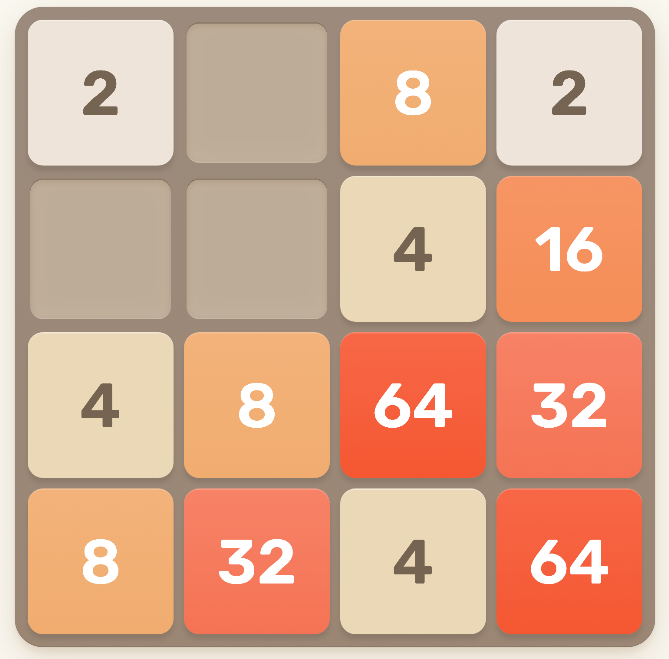}
        \caption{2048}
        \label{fig:2048}
    \end{subfigure}
    \caption{Environments for evaluation.}
    \label{fig:2048andcheetah}
\end{figure}

We evaluate \ac{ddt} in Hopper, Walker2d, and HalfCheetah (Fig. \ref{fig:cheetah}, from \cite{tassa2018deepmind}) with dense rewards.
The datasets for each environment consist of three types: medium (generated by a medium-level policy), medium-replay (sampled from the replay buffer of an agent trained to a medium performance level), and medium-expert ($1:1$ mixture of medium and expert trajectories). All datasets are obtained from the official D4RL benchmark implementation, and all reported scores are normalized using the standard evaluation interface provided by the D4RL library.

To validate the applicability of \ac{ddt} on discrete and highly stochastic problems, we employed the game 2048, a sliding puzzle game (Fig. \ref{fig:2048}, from \cite{play2048page}) where tiles with identical numbers can be combined to form a tile of greater value. 
In this environment, the agent receives a reward of $1$ only when a tile of value $128$ is created, and no reward otherwise. The offline dataset for this task consists of $5\times 10^6$ steps collected using a mixture of a random agent and an expert policy trained with PPO, as adopted from the experiments in ESPER \cite{paster2022you}.

\subsubsection{Baselines}
The baseline algorithms considered in our study fall into two categories: (1) classic offline \ac{rl} and supervised learning approaches, including BRAC-v \cite{wu2019behavior}, TD3+BC \cite{fujimoto2021minimalist}, IQL \cite{kostrikov2021offline}, and CQL \cite{kumar2020conservative}; and (2) \ac{dt} \cite{chen2021decision} and its recent variants, namely VDT \cite{zheng2025value} and LSDT \cite{wang2025long}. 
The performance scores presented in Tab. \ref{tab:mainexp} are sourced from the best results reported in the original publications or from our own implementations using d3rlpy with consistent hyperparameters, ensuring a fair comparison.

\subsubsection{Our DDT Implementation}
Our implementation builds upon d3rlpy \cite{seno2022d3rlpy}, which provides a well-structured interface for implementing and extending various offline \ac{rl} algorithms, including the original \ac{dt}.
Both \ac{ddt} and \ac{dt} employ the same GPT architecture \cite{radford2018improving} as their network backbone. Their key distinction lies in the action prediction stage. 
\ac{dt} takes a sequence that includes \ac{rtg}, $(\hat{R}_{t-k+1},o_{t-k+1},a_{t-k+1},\dots,\hat{R}_t,o_t)$, as input to the GPT and uses the final token of the output sequence for action prediction. 
In contrast, \ac{ddt} inputs a sequence without \ac{rtg}, $(o_{t-k+1},a_{t-k+1},\dots,o_t)$, into the same GPT. 
It then takes the final output token and conditions it on the return-to-go $\hat{R}_t$ via an \ac{adaln} module before predicting the action.

We adopt a popular adaLN-Zero variant to implement our \ac{ddt}, which initializes the weights of the final linear layer to zero (so that initially, $\gamma(z)=\beta(z)=0$). Specifically, in Eq. \ref{eq:adaln}, $\gamma(z)$ is replaced with $\gamma(z)+1$. This design ensures that the network behaves similarly to a standard \ac{ln} layer at the beginning of training, thereby stabilizing the training process.

The comparative experimental results with other offline \ac{rl} methods are presented in Tab. \ref{tab:mainexp}.

\begin{table*}[h]
    \centering
    \begin{tabular}{lrrrrrrrr}
        \toprule
        \textbf{Dataset \& Environment} & \textbf{BRAC-v} & \textbf{TD3+BC} & \textbf{IQL} & \textbf{CQL} & \textbf{DT} & \textbf{VDT} & \textbf{LSDT} & \textbf{DDT (Ours)} \\
        \midrule
        hopper-medium-replay-v2 & $0.6$ & $60.9$ & $94.7$ & $48.6$ & $83.1\pm0.7$ & $\mathbf{96.0}$ & $93.9$ & $92.5\pm0.6$ \\
        walker-medium-replay-v2 & $0.9$ & $81.8$ & $73.9$ & $26.7$ & $66.9\pm1.4$ & $\mathbf{82.3}$ & $74.7$ & $77.6\pm1.1$ \\
        halfcheetah-medium-replay-v2 & $\mathbf{47.7}$ & $44.6$ & $44.2$ & $46.2$ & $36.2\pm1.2$ & $39.4$ & $42.9$ & $37.8\pm0.5$ \\
        \midrule
        hopper-medium-v2 & $31.1$ & $59.3$ & $66.3$ & $58.0$ & $68.3\pm0.9$ & $98.3$ & $87.2$ & $\mathbf{99.4}\pm0.6$ \\
        walker-medium-v2 & $81.1$ & $\mathbf{83.7}$ & $78.3$ & $79.2$ & $74.3\pm0.3$ & $81.6$ & $81.0$ & $78.6\pm0.3$ \\
        halfcheetah-medium-v2 & $46.3$ & $\mathbf{48.3}$ & $47.4$ & $44.4$ & $42.5\pm0.5$ & $43.9$ & $43.6$ & $43.0\pm0.4$ \\
        \midrule
        hopper-medium-expert-v2 & $0.8$ & $98.0$ & $91.5$ & $111.0$ & $106.8\pm0.4$ & $111.5$ & $\mathbf{111.7}$ & $111.1\pm0.2$ \\
        walker-medium-expert-v2 & $81.6$ & $110.1$ & $109.6$ & $98.7$ & $108.6\pm0.3$ & $\mathbf{110.4}$ & $109.8$ & $109.5\pm0.5$ \\
        halfcheetah-medium-expert-v2 & $41.9$ & $90.7$ & $86.7$ & $62.4$ & $88.1\pm0.4$ & $93.9$ & $93.2$ & $\mathbf{94.2}\pm0.2$ \\
        \bottomrule
    \end{tabular}
    \caption{Performance of DDT and SOTA baselines on D4RL tasks. For DDT, results are reported as the mean and standard error of normalized rewards over $4\times100$ rollouts ($4$ independently trained models using different seeds with $100$ trajectories), generally showing low variance.}
    \label{tab:mainexp}
\end{table*}

\subsection{Why Does DDT Perform Better?}
\label{subsec:attention}
The core of the Transformer architecture lies in its attention module. Therefore, when analyzing methods related to \ac{dt}, visualizing the attention scores can provide a summary of how the model internally models temporal dependencies.
Given that \ac{ddt} achieves an improvement of approximately $10\%$ over \ac{dt} in the tasks listed in Tab. \ref{tab:mainexp}, we hypothesize that this improvement may be related to \ac{ddt}'s ability to allocate attention scores more efficiently and in a manner that better aligns with the inherent characteristics of these environments. The visualizations in this section present the evidence we provide in support of this hypothesis.
\begin{figure}[t]
    \centering
    \begin{subfigure}[b]{0.48\columnwidth} 
        \centering
        \includegraphics[width=\textwidth]{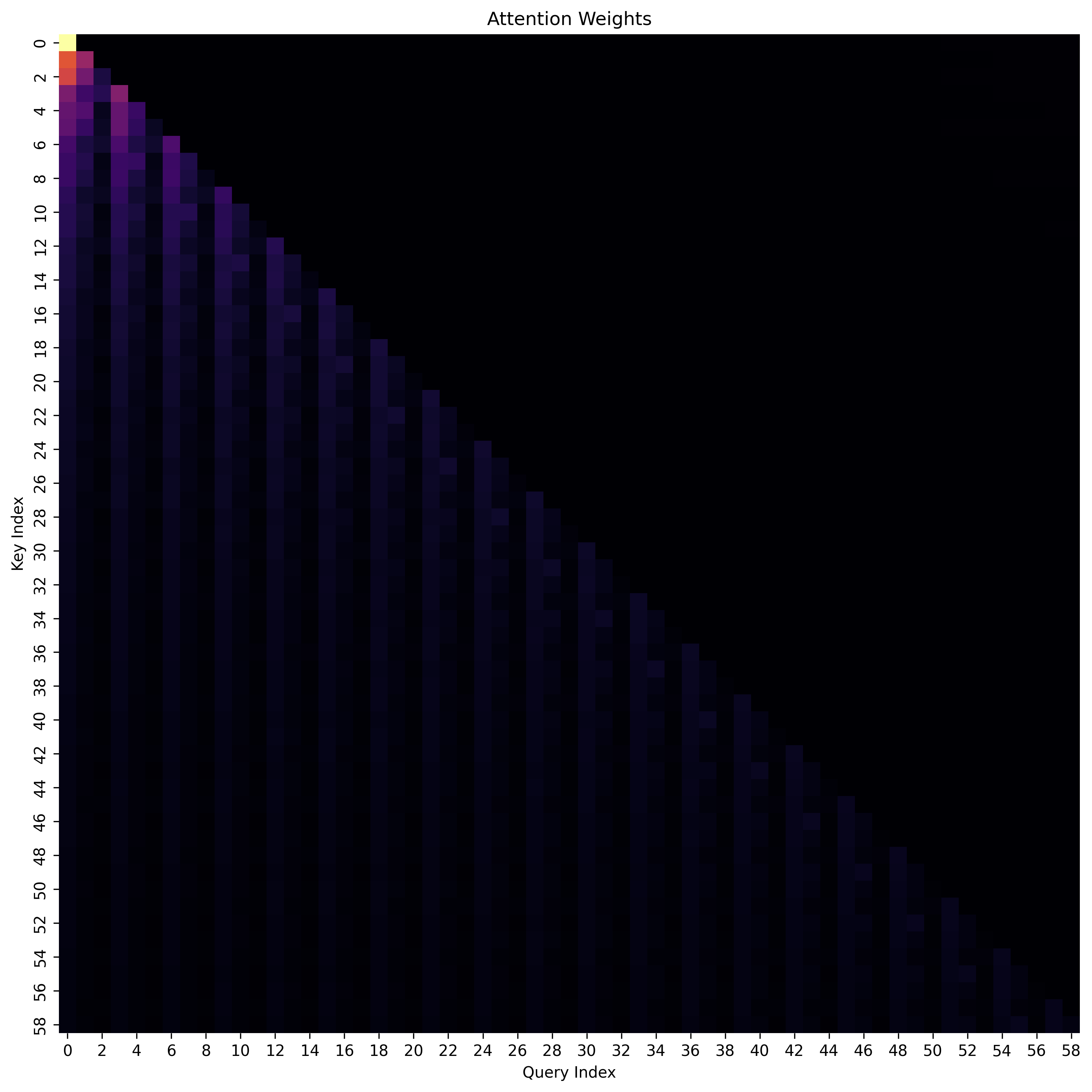}
        \caption{hopper-medium, DT}
        \label{fig:hmaphopdt}
    \end{subfigure}
    \hfill 
    \begin{subfigure}[b]{0.48\columnwidth}
        \centering
        \includegraphics[width=\textwidth]{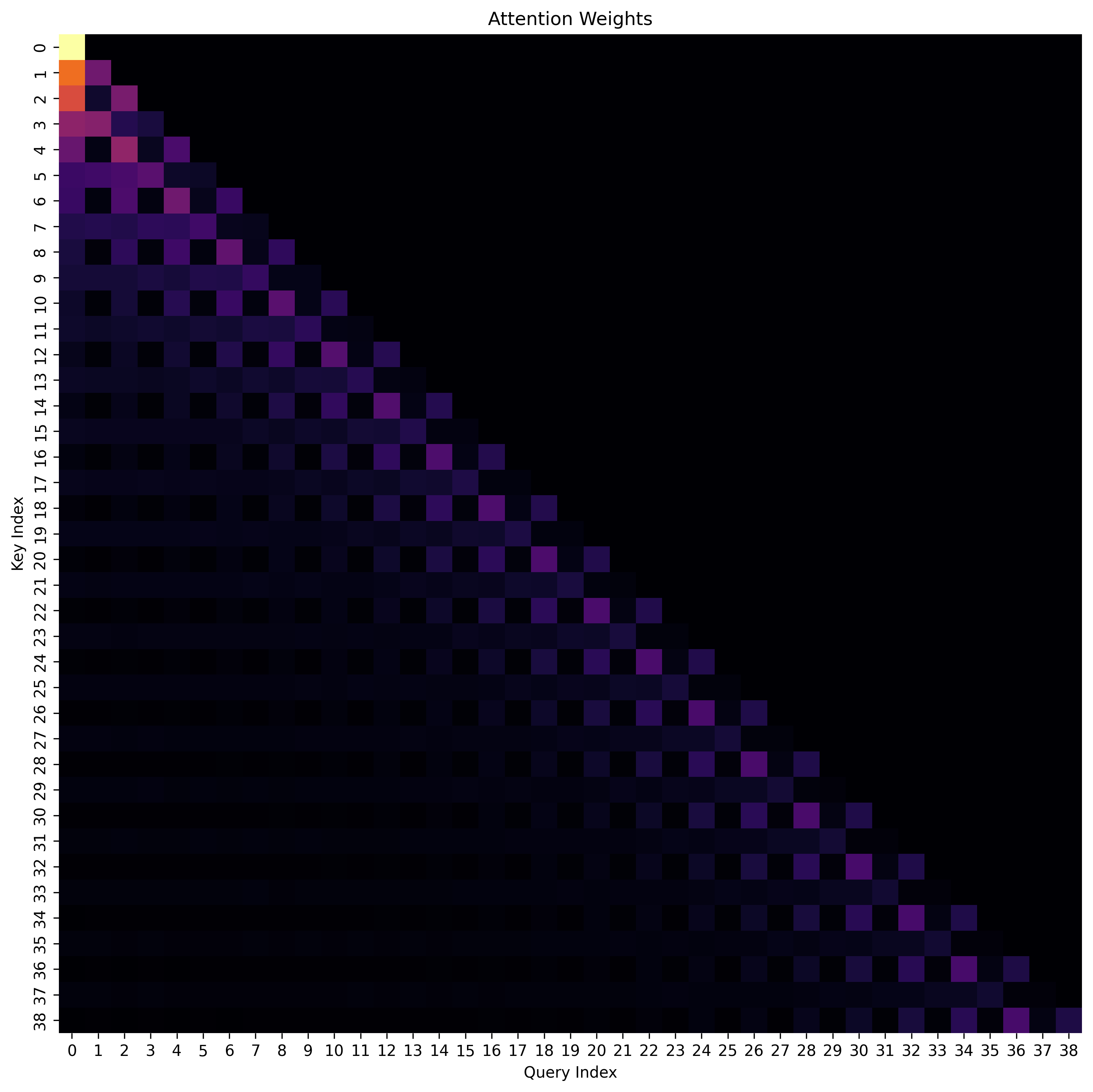}
        \caption{hopper-medium, DDT}
        \label{fig:hmaphopddt}
    \end{subfigure}
    \vspace{0.5\baselineskip} 
    \begin{subfigure}[b]{0.48\columnwidth}
        \centering
        \includegraphics[width=\textwidth]{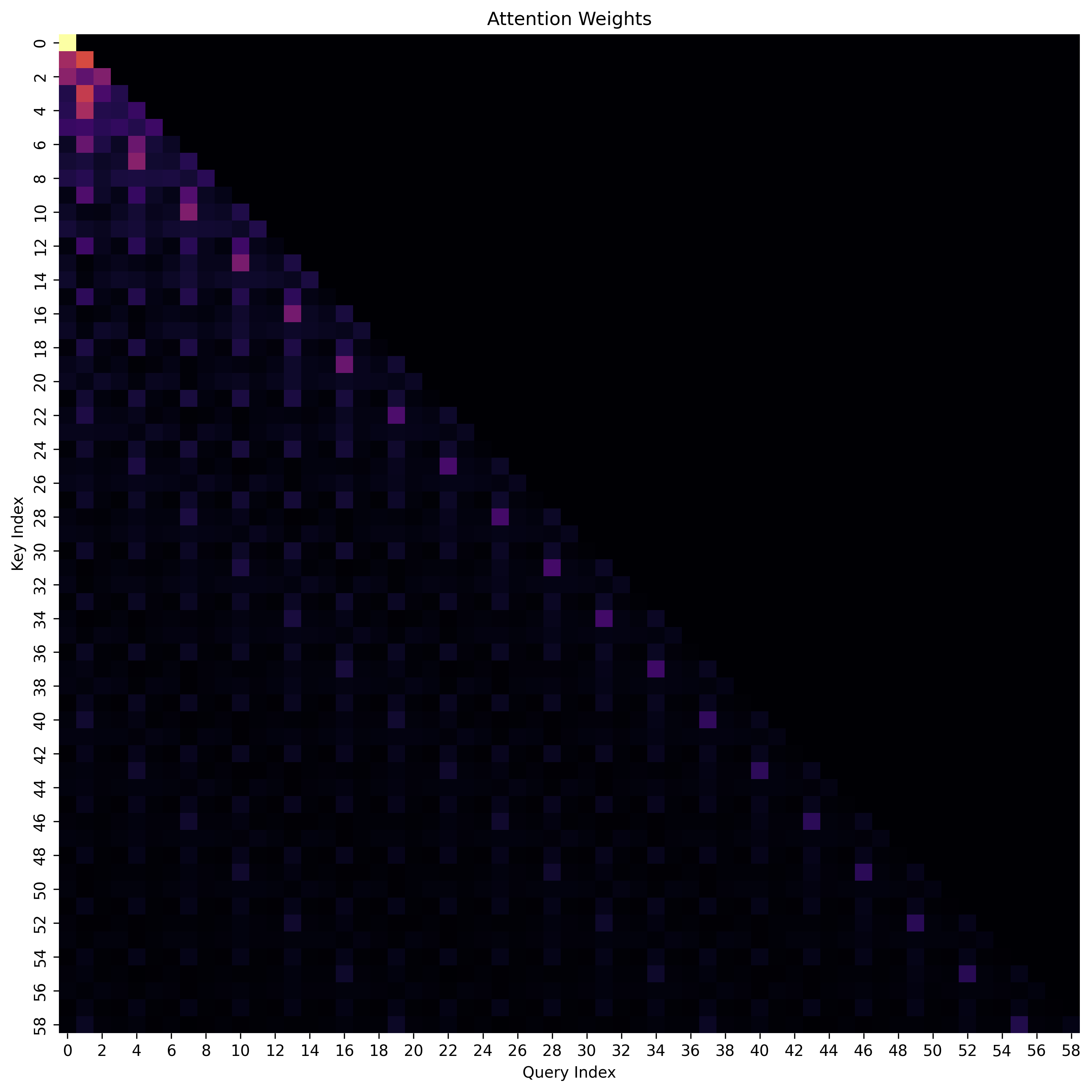}
        \caption{halfcheetah-medium-expert, DT}
        \label{fig:hmaphaldt}
    \end{subfigure}
    \hfill
    \begin{subfigure}[b]{0.48\columnwidth}
        \centering
        \includegraphics[width=\textwidth]{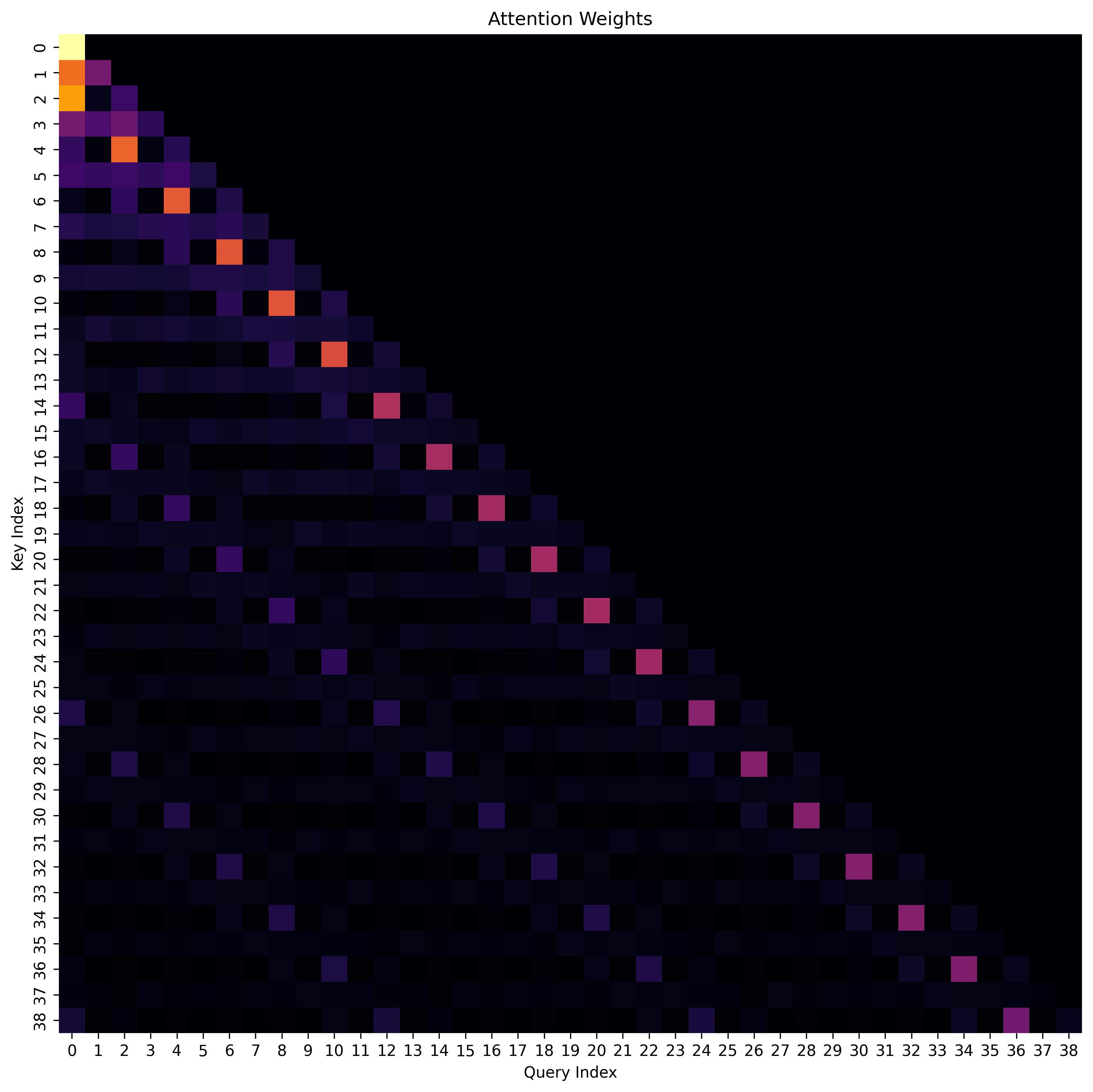}
        \caption{halfcheetah-medium-expert, DDT}
        \label{fig:hmaphalddt}
    \end{subfigure}
    \caption{Average attention scores over the first $10^3$ inference steps for the learned DT and DDT policies during action prediction in the respective environment (hopper and halfcheetah).}
    \label{fig:hmaps}
\end{figure}

The attention matrix in Fig. \ref{fig:hmaps} demonstrates the captured associations during action prediction.
These are lower triangular matrices because DT can only rely on historical information to predict actions, with all future information masked out to ensure this causality.
The horizontal and vertical axes in Fig. \ref{fig:hmaphopdt}, \ref{fig:hmaphaldt} (\ac{dt}) correspond with $(\hat{R}_1,o_1,a_1,\dots,\hat{R}_{30},o_{30},a_{30})$, and those of Fig. \ref{fig:hmaphopddt}, \ref{fig:hmaphalddt} (\ac{ddt}) correspond with $(o_1,a_1,\dots,o_{30},a_{30})$.

Since the hopper and halfcheetah environments are Markovian, theoretically, only the most recent state is required to predict actions. 
Therefore, ideally, the attention scores allocated in these environments should concentrate near the main diagonal, implying that the attention layer ought to focus more on information from adjacent timesteps. 
However, as can be observed from Fig. \ref{fig:hmaphopdt} and \ref{fig:hmaphaldt}, the attention scores learned by \ac{dt} do not exhibit such an ideal distribution; instead, they appear relatively scattered, which is also consistent with findings in existing literature \cite{wang2025long}.

As shown in Fig. \ref{fig:hmaphopddt} and \ref{fig:hmaphalddt}, the attention score distribution in \ac{ddt} aligns more closely with the ideal pattern. 
By removing the unnecessary \ac{rtg} sequences, we reduce the input length to the Transformer and mitigate the interference of such redundant information on model learning, thereby allowing attention scores to be allocated more efficiently and effectively capturing short-term dependencies.
This achieves an effectiveness analogous to that of LSDT \cite{wang2025long}, which incorporates convolution to enhance local dependency modeling.

\subsection{Does Masking Out Redundant RTGs Suffice?}
\label{subsec:maskout}
Our \ac{ddt} proposes a solution to the redundant \ac{rtg}s, as identified in Section \ref{sec:theory}.
However, in pursuit of minimizing alterations to the original \ac{dt} architecture, this is not the most concise variant to address the issue. A more streamlined form can satisfy the formulation in Eq. \ref{eq:pi} without introducing the adaLN module, merely requiring an adjustment to the causal mask to block all \ac{rtg} tokens except $\hat{R}_t$.
Here, we contrast the performance of our method with that of blocked-\ac{dt}, a variant that merely masks non-last \ac{rtg} tokens.

As shown in Tab. \ref{tab:blockdt}, our findings indicate that while blocked-\ac{dt} might achieve a modest gain over \ac{dt}, it is still inferior to \ac{ddt}. This performance gap demonstrates that for the last \ac{rtg} $\hat{R}_t$ to effectively guide the action generation, masking out redundant \ac{rtg}s $\hat{R}_{t-k+1:t-1}$ is insufficient compared to using an effective conditioning module like \ac{adaln}.

\begin{table}[h]
    \centering
    \begin{tabular}{lrrr}
        \toprule
        \textbf{Data \& Env} & \textbf{DT} & \textbf{blocked-DT} & \textbf{DDT (Ours)} \\
        \midrule
        hop-mr & $83.1$ & $82.7(-0.4)$ & $\mathbf{92.5}(+9.4)$ \\
        wlk-mr & $66.9$ & $67.2(+0.3)$ & $\mathbf{77.6}(+10.7)$ \\
        hal-mr & $36.2$ & $36.6(+0.4)$ & $\mathbf{37.8}(+1.6)$ \\
        \midrule
        hop-m & $68.3$ & $69.1(+0.8)$ & $\mathbf{99.4}(+31.1)$ \\
        wlk-m & $74.3$ & $74.1(-0.2)$ & $\mathbf{78.6}(+4.3)$ \\
        hal-m & $42.5$ & $\mathbf{43.2}(+0.7)$ & $43.0(+0.5)$ \\
        \midrule
        hop-me & $106.8$ & $107.7(+0.9)$ & $\mathbf{111.1}(+4.3)$ \\
        wlk-me & $108.6$ & $107.9(-0.7)$ & $\mathbf{109.5}(+0.9)$ \\
        hal-me & $88.1$ & $89.6(+1.5)$ & $\mathbf{94.2}(+6.1)$  \\
        \bottomrule
    \end{tabular}
    \caption{Performance comparison of DT, blocked-DT, and DDT. Numbers in parentheses indicate the performance difference relative to DT.}
    \label{tab:blockdt}
\end{table}

\subsection{Does DDT Generalize to Discrete Problems?}
\label{subsec:2048}
For discrete action spaces, we adopt the \ac{dt} implementation from d3rlpy \cite{seno2022d3rlpy} and build our \ac{ddt} variant on top of it. 
This discrete \ac{dt} differs from the standard \ac{dt} in three main aspects: (1) it uses a cross-entropy loss, (2) its output is a categorical distribution, and (3) actions are generated by selecting the one with the highest probability.
\begin{table}[h]
    \centering
    \begin{tabular}{lrrr}
        \toprule
        \textbf{Data \& Env} & \textbf{DT} & \textbf{blocked-DT} & \textbf{DDT (Ours)} \\
        \midrule
        2048 & $0.93\pm0.03$ & $0.93\pm0.02$ & $0.93\pm0.02$ \\
        \bottomrule
    \end{tabular}
    \caption{Performance of DT, blocked-DT, and DDT on 2048.}
    \label{tab:exp2048}
\end{table}

Tab. \ref{tab:exp2048} shows that \ac{ddt}, \ac{dt} and blocked-\ac{dt} achieve similar performance on 2048. 
This shows that \ac{ddt} generalizes to discrete-action, high-stochasticity, sparse-reward tasks without performance degradation, thereby establishing its applicability beyond the continuous, relatively deterministic, dense-reward control problem settings in D4RL.

\subsection{Does Adding More Layers to adaLN Help?}
\label{subsec:addlayer}
Since the conditioning variable $\hat{R}_t$ is a simple scalar, our implementation of \ac{ddt} employs a single linear layer (input dimension $1$, output dimension $128\times2$, corresponding to $\gamma$ and $\beta$ in Eq. \ref{eq:adaln}) without an activation function. 
To investigate whether a more complex \ac{adaln} could enhance performance, we evaluate a variant that uses a 2-layer \ac{mlp} with a ReLU activation in between. However, as Tab. \ref{tab:expadaln} shows on several datasets, this added complexity leads to performance degradation rather than improvement.

\begin{table}[h]
    \centering
    \begin{tabular}{lrrr}
        \toprule
        \textbf{Data \& Env} & \textbf{DT} & \textbf{DDT} & \textbf{DDT(2)} \\
        \midrule
        hop-med & $68.3$ & $99.4$ & $90.3$ \\
        hal-med & $42.5$ & $43.0$ & $42.9$ \\
        \bottomrule
    \end{tabular}
    \caption{Performance comparison. DDT(2) denotes the variant with an additional linear layer ($128 \to128$) and a ReLU in the adaLN.}
    \label{tab:expadaln}
\end{table}

\section{Further Discussion}
\label{sec:discuss}
\subsection{DDT with Generalized Condition}
Like \ac{dt}, our \ac{ddt} falls under the \ac{rcsl} paradigm \cite{kumar2019reward,brandfonbrener2022does} within the \ac{him} \cite{furutageneralized} framework. The \ac{him} offers a general theoretical framework for modeling decision-making processes that leverage both historical information and predicted future outcomes.
Beyond the \ac{rcsl} paradigm, the \ac{him} also encompasses the \ac{gcsl} paradigm. In \ac{gcsl}, the policy is conditioned on goals rather than \ac{rtg}s.

\ac{dt} generalizes effectively to \ac{gcsl} tasks by taking the entire goal-conditioned sequence as input; the following discussion explores how \ac{ddt} can achieve a similar effect under the \ac{gcsl} paradigm.

We first consider a class of scenarios where the goal remains constant throughout an episode. 
In this case, conditioning on a single goal is obviously more reasonable than conditioning on a constant sequence.
This is analogous to the case of sparse rewards in \ac{rcsl}: since the \ac{rtg} $\hat{R}_t$ stays unchanged until the episode ends, this constant \ac{rtg} can be treated as the goal. 
Our experiments on 2048 also demonstrate that \ac{ddt} can adapt effectively to this setting.

We next consider a class of scenarios where the goal may change within a single episode during the decision-making process.
The key question is whether an architecture like \ac{ddt}, which conditions only on the final goal variable, would fail in such a setting.

In \ac{gcsl}, the agent typically employs a goal-setting mechanism. 
Similar to the process in \ac{rcsl}, where a high initial \ac{rtg} $\hat{R}_1$ is specified and then updated by subtracting received rewards, the goal can be updated progressively. 
As analyzed in Section \ref{sec:theory}, in such an update scheme, past \ac{rtg}s $\hat{R}_{t-k+1:t-1}$ become redundant for current decision-making.

This scheme has a critical characteristic: the generation of the new condition (goal or \ac{rtg}) depends entirely on the observable history, including observations, actions, and rewards. 
This information is either directly fed into the sequence model or can be reliably inferred by it (e.g., in a \ac{pomdp}, the reward sequence can be deduced from the observation-action sequence, providing no additional decision-relevant information). 
In such cases, the \ac{ddt} architecture remains effective. It does not need to rely on the full sequence of redundant goal conditions; conditioning on the latest goal is sufficient for action generation.

However, if the goal update mechanism in \ac{gcsl} does not possess this characteristic, for instance, if the goal at each step depends on information external to the observable history, such as being specified by an oracle with an unknown internal state and transition dynamics, then our \ac{ddt} framework would no longer be applicable. 
In this scenario, providing the entire sequence of past goals to the model would convey strictly more information than providing only the latest one.
One possible solution to this problem lies in augmenting the observations provided to the model so that they encompass the information internal to the oracle.

\subsection{Role of RTG Sequence in DT}

Our analysis in Section \ref{sec:theory} demonstrates that all but the final item in the \ac{rtg} sequence are redundant. 
While this observation might appear to suggest that paradigms like \ac{dt}, which utilize the entire \ac{rtg} sequence as input, are unnecessary, our findings indicate that such paradigms nevertheless hold value under certain circumstances, despite might not being the most efficient implementation.

One scenario involves problems that do not conform to the standard \ac{pomdp} assumption. In such cases, updating the belief state relies not only on the observation and action sequences but also on the reward. Here, considering the entire \ac{rtg} sequence effectively encodes the reward information from each step, which can provide gains for decision-making.

The second scenario involves standard \ac{pomdp}s. Although the \ac{rtg} sequence is theoretically redundant, it can improve learning by enriching the feature representation. A typical case is when states with minimal visual differences yield vastly different immediate rewards. Providing per-step rewards enables the network to discern these critical subtleties and better predict high-return actions.

While the \ac{rtg} sequence can provide gains in the aforementioned two scenarios, we hold a different perspective from \ac{dt} regarding how to utilize the per-step reward information it contains.
We posit that when per-step reward or \ac{rtg} information is necessary as model input, it should be incorporated as a field within the observation, rather than being treated as an independent input token to the Transformer, parallel to observations and actions. This preference is based on two considerations: From the perspective of computational cost, the quadratic complexity of the Transformer's input sequence favors shorter sequences. In terms of input dimensions and information content, per-step reward is a scalar that carries significantly less information compared to the higher-dimensional observation and action vectors, making it unsuitable as an independent token. 
Even in traditional RNN models, shorter sequences still play a crucial role in enabling efficient policy learning.


\section{Conclusion and Future Work}
\label{sec:conclude} 
In this work, we identify and address a fundamental redundancy in the \ac{dt} framework: conditioning the policy on the full history of \ac{rtg}s is theoretically unnecessary and empirically harmful to performance, an issue previously overlooked in the literature. Our proposed \ac{ddt} architecture resolves this by utilizing \ac{adaln} to condition solely on the current \ac{rtg}. This principled simplification yields a model that is not only more computationally efficient but also achieves superior performance, establishing a strong and efficient baseline.

Future work includes several promising directions:

(1) Advanced Condition Fusion: Exploring the application of \ac{dit}'s sophisticated, layer-wise condition modulation paradigms to \ac{dt} architectures.

(2) Architectural Generalization: Applying the \ac{ddt} paradigm, conditioning on minimal necessary information, to other \ac{dt} variants to enhance their efficiency and performance.
\bibliographystyle{named}
\bibliography{ijcai26}

\end{document}